\documentclass[discussion]{clv3}

\usepackage{hyperref}
 \usepackage{soul}
\usepackage{xcolor}
\definecolor{darkblue}{rgb}{0, 0, 0.5}
\hypersetup{colorlinks=true,citecolor=darkblue, linkcolor=darkblue, urlcolor=darkblue}

\usepackage{amsmath}
\usepackage{framed}
\usepackage{booktabs}
\usepackage{cleveref}
\usepackage{subcaption}
\usepackage[compact]{titlesec}

\issue{1}{1}{2016}

\dochead{Squib}

\runningtitle{Probing Classifiers}

\runningauthor{Yonatan Belinkov}

\begin{document}

\title{Probing Classifiers: Promises, Shortcomings, and Advances}

\historydates{Submission received: 4 March 2021; 
             revised version received: 31 July 2021; 
             accepted for publication:  8 September 2021}

\author{Yonatan Belinkov\thanks{Supported by the Viterbi Fellowship in the Center for Computer Engineering at the Technion.}}
\affil{Technion -- Israel Institute of Technology \\ {\tt belinkov@technion.ac.il}}




\maketitle

\begin{abstract}
Probing classifiers have emerged as one of the prominent methodologies for interpreting and analyzing deep neural network models of natural language processing. The basic idea is simple --- a classifier is trained to predict some linguistic property from a model's representations --- and has been used to examine a wide variety of models and properties. However, recent studies have demonstrated various methodological limitations of this approach. This article critically reviews the probing classifiers framework, highlighting their promises, shortcomings, and advances. 
\end{abstract}

\section{Introduction}

\looseness=-1
The opaqueness of deep neural network models of natural language processing (NLP) has spurred a line of research into interpreting and analyzing them. 
Analysis methods may aim to answer questions about a model's structure or its decisions. For instance, one might  ask which parts of a neural neural model are responsible for certain linguistic properties, or which parts of the input led the model to make a certain decision. 
A common methodology to answer questions about the structure of models is to associate internal representations with external properties, by training a classifier on said representations that predicts a given property. This framework, known as \mbox{\textbf{probing classifiers}}, has emerged as a prominent analysis strategy in many studies of NLP models.\footnote{For an overviews of analysis methods in NLP, see the survey by \citet{belinkov-glass-2019-analysis}, as well as the tutorials by \citet{belinkov-etal-2020-interpretability} and \citet{wallace-etal-2020-interpreting}. For an overview of explanation methods in particular, see the survey by \citet{danilevsky-etal-2020-survey}.}  

\looseness=-1
Despite its apparent success, the probing classifiers paradigm is not without limitations. Critiques have been made about comparative baselines, metrics, the choice of classifier, and the correlational nature of the method. In this short article, we first define the probing classifiers framework, taking care to consider the various involved components. Then we summarize the framework's shortcomings, as well as improvements and advances. 
This article provides a roadmap for NLP researchers who wish to examine probing classifiers more critically and highlights areas in need of additional research.

\section{The Probing Classifiers Framework} \label{sec:framework}

On the surface, the probing classifiers idea seems straightforward. We take a model that was trained on some task, such as a language model. We generate representations using the model, and train another classifier that takes the representations and predicts some property. If the classifier performs well, we say that the model has learned information relevant for the property. 
However, upon closer inspection, it turns out that much more is involved here. To see this, we now define this framework a bit more formally. 

Let us denote by $f : x \mapsto \hat{y}$ a model that maps input $x$ to output $\hat{y}$. We call this model the original model. It is trained on some annotated dataset $\mathcal{D}_O = \{x^{(i)}, y^{(i)}\}$, which we refer to as the original dataset. Its performance is evaluated by some measure, denoted $\textsc{Perf}(f, \mathcal{D}_O)$.
The function $f$ is typically a deep neural network that generates intermediate representations of $x$, for example $f_l(x)$ may denote the representation of $x$ at layer $l$ of $f$.\footnote{We use $f_l(x)$ to refer more generally to any intermediate output of $f$ when applied to $x$, so the framework includes analyses of other model components, such as attention weights \cite{clark-etal-2019-bert}.} 
A probing classifier $g : f_l(x) \mapsto \hat{z}$ maps intermediate representations to some property $\hat{z}$, which is typically some linguistic feature of interest. 
As a concrete example, $f$ might be a sentiment analysis model, mapping a text $x$ to a sentiment label  $y$, while $g$ might be a classifier mapping intermediate representations $f_l(x)$ to part-of-speech tags $z$.
The classifier $g$ is trained and evaluated on some annotated dataset $\mathcal{D}_P = \{x^{(i)}, z^{(i)}\}$, and some performance measure $\textsc{Perf}(g, f, \mathcal{D}_O, \mathcal{D}_P)$ (e.g., accuracy) is reported. Note that the performance measure depends on the probing classifier $g$ and the probing dataset $\mathcal{D}_P$, as well as on the original model $f$ and the original dataset $\mathcal{D_O}$.

From an information theoretic perspective, training the probing classifier $g$ can be seen as estimating the mutual information between the intermediate representations $f_l(x)$ and the property $z$ (\citealt[p. 42]{belinkov:2018:phdthesis}; \citealt{pimentel-etal-2020-information}; \citealt{zhu-rudzicz-2020-information}), which we write $\mathrm{I}(\mathbf{z} ; \mathbf{h})$, where  $\mathbf{z}$ is a random variable ranging over properties $z$ and $\mathbf{h}$ is a random variable ranging over representations $f_l(x)$.  

The above careful definition of the probing classifiers framework reveals that it is comprised of multiple concepts and components, depicted in \Cref{fig:probing-components-basic}.  The choice of each such component, and the interactions between them, lead to non-trivial questions regarding the design and implementation of any probing classifier experiment. Before we turn to these considerations   in \Cref{sec:shortcomings-advances}, we briefly review some history and promises of probing classifiers in the next section.

\begin{figure}[h]
    \centering
    \begin{subfigure}[b]{\textwidth}
    \centering
        \begin{tabular}{l @{\hskip 1em} l } 
        \toprule
         $x \mapsto y$ & Original task \\
         $\mathcal{D}_O = \{x^{(i)}, y^{(i)}\} $ & Original dataset \\
         $f : x \mapsto y $ & Original model \\
         $\textsc{Perf}(f, \mathcal{D}_O)$ & Performance on the original task \\
         $f_l(x)$ & Representations of $x$ from $f$\\
         $f_l(x) \mapsto z$ & Probing task \\
         $\mathcal{D}_P = \{x^{(i)}, z^{(i)}\} $ & Probing dataset \\
         $g : f_l(x) \mapsto z$ & Probing classifier \\
         $\textsc{Perf}(g, f, \mathcal{D}_O, \mathcal{D}_P) $ & Probing performance  \\ 
         \bottomrule
        \end{tabular}
         \caption{Basic Components.}
         \label{fig:probing-components-basic}
     \end{subfigure}
     \begin{subfigure}[b]{\textwidth}
     \centering
        \begin{tabular}{l @{\hskip 1em} l } 
        \toprule          
         $\bar{f} : x \mapsto y$ & Skyline model or upper bound \\ 
         $\underline{f} : x \mapsto y$ & Baseline model \\          
         $x \mapsto y_{Rand}$ & Control task \cite{hewitt-liang-2019-designing} \\ 
         $c : f_l(x) \mapsto c(f_l(x)) $ & Control function \cite{pimentel-etal-2020-information} \\ 
         $\mathcal{D}_{P,Rand}$ & Control task dataset \cite{hewitt-liang-2019-designing} \\ 
         $\mathcal{D}_{O,z}$ & Control dataset \cite{ravichander:2021:eacl} \\          
         $\textsc{Sel}(g, f, \mathcal{D}_O, \mathcal{D}_P, \mathcal{D}_{P,Rand})$ & Probing selectivity \cite{hewitt-liang-2019-designing} \\
         $ \mathcal{G}(\mathbf{z}, \mathbf{h}, c) $ & Information gain w.r.t control function \cite{pimentel-etal-2020-information} \\ 
         $\textsc{MDL}(g, f, \mathcal{D}_O, \mathcal{D}_P)$ & Probe minimum description length \cite{voita-titov-2020-information} \\ 
         $\tilde{f}_l(x)$ & Representations of $x$ from $f$, after an intervention \\ 
         \bottomrule 
        \end{tabular}
        \caption{Additional Components.}
        \label{fig:probing-components-extended}
        \vspace{-3pt}
    \end{subfigure}
    \caption{Components comprising the probing classifiers framework.}
    \label{fig:probing-components}
    \vspace{-19pt}
\end{figure}

\section{Promises} \label{sec:promises}

\looseness=-1
Perhaps the first studies that can be cast in the framework of probing classifiers are by \citet{kohn-2015-whats} and \citet{gupta-etal-2015-distributional}, who trained classifiers on static word embeddings to predict various morphological, syntactic, and semantic properties. Their goals were to provide more nuanced evaluations of word embeddings compared to prior work, which only integrated them in downstream tasks.  
Other early work classified hidden states of a recurrent neural network machine translation system into morpho-syntactic properties \cite{shi-etal-2016-string}. They were motivated by the end-to-end nature of the neural machine translation system, which, compared to a phrase/syntax-based system, did not explicitly integrate such properties (so they ask: ``What kind of syntactic information is learned, and how much?'').  
The framework has taken up a more stable form by several groups who studied sentence embeddings \cite{ettinger-etal-2016-probing,adi:2017:ICLR,conneau-etal-2018-cram}  and recurrent/recursive neural networks \cite{belinkov-etal-2017-neural,hupkes2018visualisation}.\footnote{For chronological completeness, workshop and preprint versions of \citet{hupkes2018visualisation} and \citet{adi:2017:ICLR} appeared earlier \cite{veldhoen2016diagnostic,DBLP:journals/corr/AdiKBLG16}.}  The same idea had been concurrently proposed for investigating computer vision models \cite{alain2016understanding}.

A main motivation in this body of work is the \emph{opacity} of the representations.\footnote{``little is known about the information that is captured by different sentence embedding learning mechanisms'' \cite{adi:2017:ICLR}; ``a poor understanding of what they are capturing'' \cite{conneau-etal-2018-cram}; ``little is known about what and how much these models learn.'' 
\cite{belinkov-etal-2017-neural}.} 
Compared to performance on downstream tasks, probing classifiers aim to provide more nuanced evaluations w.r.t \emph{simple properties}.\footnote{``fine-grained measurement of some of the information encoded in sentence embeddings'' \cite{adi:2017:ICLR}; ``simple linguistic properties of sentences'' \cite{conneau-etal-2018-cram}; ``assessing the specific semantic information that is being captured in sentence representations'' \cite{ettinger-etal-2016-probing}.} 
Indeed, following the initial studies, a plethora of work has applied the framework to various models and properties, alleviating some of the opacity, at least in terms of properties encoded in the representations. See \citet{belinkov-glass-2019-analysis} for a comprehensive survey up to early 2019.\footnote{There have also been numerous other studies using the probing classifier framework as is. For a partial list, see \url{https://github.com/boknilev/nlp-analysis-methods/issues/5}. For recent analyses focusing on the BERT model \cite{devlin-etal-2019-bert}, see  \citet{rogers-etal-2020-primer}.}  

However, what can be inferred from successful probing performance is less obvious. 
Good probing performance is often taken to indicate several potential situations: 
good  \emph{quality} of the representations w.r.t the probing property,\footnote{``evaluate the quality of the trained classifier on the given task as a proxy to the quality of the extracted representations'' \cite{belinkov-etal-2017-neural}.}
\emph{readability} of information found in the representations,\footnote{``If the classifier succeeds, it means that the pre-trained encoder is storing readable tense information into the embeddings it creates'' \cite{conneau-etal-2018-cram}.}   
or its \emph{extractability}.\footnote{``testing for extractability of semantic information by testing classification accuracy..'' \cite{ettinger-etal-2016-probing}; ``if a sequential model is computing certain information, or merely keeping track of it, it should be possible to extract this information from its internal state space'' \cite{hupkes2018visualisation}.}
In contrast, low probing performance is taken to indicate that the probing property is not present in the representations or is not usable.\footnote{``low accuracy suggests this information is not represented in the hidden state'' \cite{hupkes2018visualisation}; ``if we cannot train a classifier to predict some property of a sentence based on its vector representation, then this property is not encoded in the representation (or rather, not encoded in a useful way, considering how the representation is likely to be used)'' \cite{adi:2017:ICLR}.} 
Sometimes, good  performance is taken to indicate \emph{how} the original model achieves its behavior on the original task \cite{hupkes2018visualisation}. A linear probing classifier is thought to reveal features that are used by the original model, while a more complex probe ``bears the risk that the classifier infers features that are not actually used by the network'' \cite{hupkes2018visualisation}.  
Often, different terms (\emph{quality}, \emph{readability}, \emph{usability}, etc.) appear abstractedly without precise definitions.

As we shall see, some of the above assumptions and conclusions are better accounted for than others by the probing classifiers paradigm. 
Indeed, the community has recently taken a more critical look at the methodology, which we turn to now.

\section{Shortcomings and Advances} \label{sec:shortcomings-advances}

In light of the promises discussed above, this section reviews several limitations of the probing classifiers framework, as well as existing proposals for addressing them. We discuss comparisons and controls, how to choose the probing classifier, which causal claims can be made, the difference between datasets and tasks, and the need to define the probed properties. 
We formalize new additional components (\Cref{fig:probing-components-extended}) in a unified framework, along with the basic components (\Cref{fig:probing-components-basic}).

\subsection{Comparisons and controls} 

A first concern with the framework is how to interpret the results of a probing classifier experiment. 
Suppose we run such an experiment and obtain a performance of $\textsc{Perf}(g, f, \mathcal{D}_O, \mathcal{D}_P) = 87.8$. Is that a high/low number? What should we compare it to? 
We will denote a baseline model with $\underline{f}$ and an upper bound or skyline model with $\bar{f}$. 

Some studies compare with majority baselines \cite{belinkov-etal-2017-neural,conneau-etal-2018-cram} or with classifiers trained on representations that are thought to be simpler than what the original model $f$ produces, such as static word embeddings \cite{belinkov-etal-2017-neural,tenney2018what}.  Others advocate for random baselines, training the classifier $g$ on a randomized version of $f$ \cite{conneau-etal-2018-cram,zhang-bowman-2018-language,tenney2018what,chrupala-etal-2020-analyzing}. These studies show that even random features capture significant information that can be decoded by the probing classifier, so performance on learned features should be viewed in such a perspective. 

On the other hand, some studies compare $\textsc{Perf}(g, f, \mathcal{D}_O, \mathcal{D}_P)$ to skylines or upper bounds $\bar{f}$, in an attempt to provide a point of comparison for how far probing performance is from the possible performance on the task of mapping $x \mapsto z$. 
Examples include estimating human performance \cite{conneau-etal-2018-cram}, reporting the state of the art from the literature \cite{liu-etal-2019-linguistic}, or training a dedicated model to predict $z$ from $x$, without restricting to (frozen) representations from $f$ \cite{belinkov-etal-2017-evaluating}.

Others have proposed to design controls for possible confounders. \citet{hewitt-liang-2019-designing} observe that the probing performance  $\textsc{Perf}(g, f, \mathcal{D}_O, \mathcal{D}_P)$ may tell us more about the probe $g$ than about the model $f$. The probe $g$ may memorize information from $\mathcal{D}_P$, rather than evaluate information found in representations $f(x)$.   They design control tasks, which a probe may only solve by memorizing. In particular, they randomize the labels in  $\mathcal{D}_P$, creating a new dataset  $\mathcal{D}_{P,Rand}$. Then, they define \emph{selectivity} as the difference between the probing performance on the probing task and the control task:  $\textsc{Sel}(g, f, \mathcal{D}_O, \mathcal{D}_P, \mathcal{D}_{P,Rand})$ =  $\textsc{Perf}(g, f, \mathcal{D}_O, \mathcal{D}_P) - \textsc{Perf}(g, f, \mathcal{D}_O, \mathcal{D}_{P,Rand})$. They show that probes may have high accuracy, but low selectivity, and that linear probes tend to have high selectivity, while non-linear probes tend to have low selectivity. This indicates that high accuracy of non-linear probes may come from memorization of surface patterns by the probe $g$, rather than from information captured in the representations $f_l(x)$. 
The control tasks introduced by \citeauthor{hewitt-liang-2019-designing} are particularly suited for word-level properties $z$ as they evaluate memorization of word types; it is less clear how to apply this idea more broadly, such as in sentence-level properties. 

Taking an information-theoretic perspective on probing, \citet{pimentel-etal-2020-information} proposed to use control functions instead of control tasks in order to compare probes. Their control function is any function applied to the representation, $c : f_l(x) \mapsto c(f_l(x))$, and they compare the information gain, which is the difference in mutual information between the property $z$ and the representation before and after applying the control function:  $ \mathcal{G}(\mathbf{z}, \mathbf{h}, c) =   \mathrm{I}(\mathbf{z} ; \mathbf{h}) - \mathrm{I}(\mathbf{z} ; \mathbf{c(h)})$. 
While \citet{pimentel-etal-2020-information} posit that their control function are a better criterion than the control tasks of \citet{hewitt-liang-2019-designing}, subsequent work showed that the two criteria are almost equivalent, both theoretically and empirically \cite{zhu-rudzicz-2020-information}.

Another kind of control is proposed by \citet{ravichander:2021:eacl}, who design control datasets, where the linguistic property $z$ is not discriminative w.r.t the original task of mapping $x$ to $y$. That is, they modify $\mathcal{D}_O$ and create a new dataset, $\mathcal{D}_{O,z}$, where all examples have the same value for property $z$. Intuitively, a model $f$ trained on $\mathcal{D}_{O,z}$ should not pick up information about $z$, since it is not useful for the task of $f$. They show that a probe $g$ may learn to predict property $z$ incidentally, even when it is not discriminative w.r.t the original task of mapping $x \mapsto y$, casting doubts on causal claims concerning the effect that a property encoded in the representation may have on the original task. While they create control datasets for probing sentence-level information, the same idea can be applied to word-level properties.

\subsection{Which classifier to use?}

Another concern is the choice of the probing classifier $g$: 
What should be its structure? What role does its expressivity play in drawing conclusions about the original model $f$? 

Some studies advocate for using simple probes, such as linear classifiers \cite{alain2016understanding,hupkes2018visualisation,liu-etal-2019-linguistic,hall-maudslay-etal-2020-tale}. Somewhat anecdotally, a few studies observed better performance with more complex probes, but reported similar relative trends \cite{conneau-etal-2018-cram,belinkov:2018:phdthesis}. That is, a ranking 
 $\textsc{Perf}(g, f_1, \mathcal{D}_O, \mathcal{D}_P) > \textsc{Perf}(g, f_2, \mathcal{D}_O, \mathcal{D}_P)$, of two representations $f_1(x)$ and $f_2(x)$,  holds across different probes $g$. 
However, this pattern may be flipped under alternative measures, such as selectivity \cite{hewitt-liang-2019-designing}. 

Several studies considered the complexity of the probe $g$ in more detail. \citet{pimentel-etal-2020-information} argue that, in order to give the best estimate about the information that model $f$ has about property $z$, the most complex probe should be used. 
In a more practical view, \citet{voita-titov-2020-information} propose to measure both the performance of the probe $g$ and its complexity, by estimating the minimum description length of the code required to transmit property $z$ knowing the representations $f_l(x)$: 
$\textsc{MDL}(g, f, \mathcal{D}_O, \mathcal{D}_P)$.
Note that this measure again depends on the probe $g$, the model $f$, and their respective datasets $\mathcal{D}_O$ and  $\mathcal{D}_P$. 
They found that MDL provides more information about how a probe $g$ works, for instance by revealing differences in complexity of probes when performing control tasks from $\mathcal{D}_{P,Rand}$, as in \citet{hewitt-liang-2019-designing}. 
 \citet{pimentel-etal-2020-pareto} argue that probing work should report the possible trade-offs between accuracy and complexity, along a range of probes $g$, and call for using probes that are both simple and accurate. 
While they study a number of linear and  non-linear multi-layered perceptrons, one could extend this idea to other classes of probes. Indeed, \citet{cao2021low} design a pruning-based probe, which learns a mask on weights of $f$ and obtains a  better accuracy--complexity trade-off than a non-linear probe.

Another line of work proposes methods to extract linguistic information from a trained model without learning additional parameters. In particular, much work has used some sort of pairwise importance score between words in a sentence as a signal for inferring linguistic properties, either full syntactic parsing or more fine-grained properties such as coreference resolution. These scores may come from attention weights \cite{raganato-tiedemann-2018-analysis,clark-etal-2019-bert,marecek-rosa-2019-balustrades,htut2019attention} or from distances between word representations, perhaps including perturbations of the input sentence \cite{wu-etal-2020-perturbed}.   The pairwise scores can feed into some general parsing algorithm, such as the Chu-Liu Edmonds algorithm \citeyearpar{10030090917,edmonds1967optimum}.  Alternatively, some work has used representational similarity analysis \cite{10.3389/neuro.06.004.2008} to measure similarity between word or sentence representations and syntactic properties, both local properties like determining a verb's subject \cite{lepori-mccoy-2020-picking} and more structured properties like inferring the full syntactic tree \cite{chrupala-alishahi-2019-correlating}. Also related is work on clustering representations w.r.t linguistic property and classifying by cluster assignment \cite{zhou-srikumar-2021}.  
This line of work can be seen as a parameter-less probing classifier $g$: a linguistic property is inferred from internal model components (representations, attention weights), without needing to learn new parameters. Thus, such work avoids some of the issues about what the probe learns. Additionally, from the perspective of an accuracy--complexity trade-off, such work should perhaps be placed on the low end of the complexity axis, although the complexity of the parsing algorithm could also be taken into account.

\subsection{Correlation vs.\ causation} \label{sec:causal}

A main limitation of the probing classifier paradigm is the disconnect between the probing classifier $g$ and the original model $f$. They are trained in two different steps, where $f$ is trained once and only used to generate feature representations $f_l(x)$, which are fed into $g$. Once we have $f_l(x)$, we get a probing performance from $g$, which tells us something about the information in  $f_l(x)$. However, in the process, we have forgotten about the original task assigned to $f$, which was to predict $y$. This raises an important question, which early work has largely taken for granted (\Cref{sec:promises}): 
Does model $f$ use the information discovered by probe $g$? 
In other words, the probing framework may indicate correlations between representations $f_l(x)$ and linguistic property $z$, but it does not tell us whether this property is involved in predictions of $f$. 
Indeed, several studies pointed out this limitation \cite{belinkov-glass-2019-analysis}, including reports on a mismatch between performance of the probe, $\textsc{Perf}(g, f, \mathcal{D}_O, \mathcal{D}_P)$, and performance of the original model, $\textsc{Perf}(f, \mathcal{D}_O)$  \cite{VanmassenhoveDuWay2017}. 
In contrast, \citet{lovering2021predicting} find that extractability of a property according to $\textsc{MDL}(g, f, \mathcal{D}_O, \mathcal{D}_P)$ is correlated with $f$ making predictions consistent with that property. 
Relatedly, \citet{tamkin-etal-2020-investigating} find a discrepancy between features $f_l(x)$ obtaining high probing performance, $\textsc{Perf}(g, f, \mathcal{D}_O, \mathcal{D}_P)$, and features identified as important when fine-tuning $f$ while performing the probing task $f_l(x) \mapsto z$. They reveal this by randomizing the weights of specific layers when fine-tuning $f$, which can be seen as a kind of intervention.

Indeed, a number of studies have proposed improvements to the probing classifier paradigm, which aim to discover causal effects by \emph{intervening} in representations of the model $f$. 
\citet{giulianelli-etal-2018-hood} use gradients from $g$ to modify the representations in $f$ and evaluate how this change affects both the probing performance and the original model performance. In their case, $f$ is a language model and $g$ predicts subject--verb number agreement. They find that their intervention increases probing performance, as may be expected. Interestingly, while in the general language modeling case the intervention has a small effect on the original model performance, $\textsc{Perf}(f, \mathcal{D}_O)$, they find an increase in this performance on examples designed to assess number agreement. They conclude that probing classifiers can identify features that are actually used by the model. 
\citet{tucker2021modified} also use probe gradients to update the representations $f_l(x)$ w.r.t $z$, resulting in what they call counterfactual representations, and measure the effect on other properties. 
Similarly, \citet{elazar2020amnesic} remove certain properties $z$ (such as parts of speech or syntactic dependencies) from representations in $f$ by repeatedly training (linear) probing classifiers $g$ and projecting them out of the representation. This results in a modified representation $\tilde{f}_l(x)$, which has less information about $z$.  They compare the probing performance to the performance on the original task (in their case, language modeling) after the removal of said features. They find that high probing performance $\textsc{Perf}(g, f, \mathcal{D}_O, \mathcal{D}_P)$ does not necessarily entail a large drop in original task performance after their removal, that is, $\textsc{Perf}(\tilde{f}, \mathcal{D}_O)$. Thus, contrary to \citet{giulianelli-etal-2018-hood}, they conclude that probing classifiers do not always identify features that are actually used by the model. 
In a similar vein, \citet{feder2020causalm} remove properties $z$ from representations in $f$ by training $g$ adversarially. 
At the same time, another probing classifier $g_C$ is trained positively, aiming to control for properties $z_C$ that should not be removed from $f$. A major difference from standard probing classifiers work is the continued updating of $f$. They find that they can accurately estimate the effect of properties $z$ on downstream tasks performed by $f$ when it is fine-tuned.\footnote{Other studies that perform interventions to interpret NLP models without involving probing classifiers \cite[e.g.,][]{bau2018identifying,lakretz-etal-2019-emergence,vig:2020:neurips} are left out of the present scope.}

\subsection{Datasets vs.\ tasks}

The probing paradigm aims to study models performing some task ($f : x \mapsto \hat{y}$) via a classifier performing another task ($g: f_l(x) \mapsto \hat{z}$). However, in practice these \emph{tasks} are operationalized via finite \emph{datsaets}. 
\citet{ravichander:2021:eacl}  point out that datasets are imperfect proxies for tasks. 
Indeed, 
the effect of the choice of datasets---both the original dataset $\mathcal{D}_O$ and the probing dataset $\mathcal{D}_P$---has not been widely studied. Furthermore, we ideally want to disentangle the role of each dataset from the role of the original model $f$ and probing classifier $g$. 
Unfortunately, models $f$ tend to be trained on different datasets $\mathcal{D}_O$, making statements about models confounded with issues of datasets. Some prior work acknowledged that conclusions can only be made about the existing \emph{trained models}, not about general \emph{architectures} \cite{liu-etal-2019-linguistic}. 
However, in an ideal world, we would compare different architectures $\{f^i\}$ trained on the same dataset $\mathcal{D}_O$  or the same  $f$ trained on different datasets $\{\mathcal{D}_O^i\}$. 
Concerning the latter, \citet{zhang-etal-2021-need} found that models require less data to encode syntactic and semantic properties compared to commonsense knowledge.
More such experiments are currently lacking.  

The effect of the probing dataset $\mathcal{D}_P$---its size, composition, etc.---is similarly not well studied. While some work reported results on multiple datasets when predicting the same property $z$ \cite[e.g.,][]{belinkov-etal-2017-neural}, more careful investigations are needed.

\subsection{Properties must be pre-defined}

Finally, 
inherent to the probing classifier framework is determining a property $z$ to probe for. This limits the investigation in multiple ways: It constrains the work to existing annotated datasets, which are often limited to English and certain properties. It also requires focusing on properties $z$ that are thought to be relevant to the task of mapping $x \mapsto y$ a-priori, potentially leading to biased conclusions.
In an isolated effort to alleviate this limitation, \citet{michael-etal-2020-asking} propose to learn latent clusters useful for predicting a property $z$. They discover clusters corresponding to known properties (such as personhood) as well as new categories, which are not usually annotated in common datasets. Still, probing classifiers are so far mainly useful when one has prior expectations about which properties $z$ might be relevant w.r.t a given task.

\section{Summary}

Given the various limitations discussed in this article, one might ask: 
What are probing classifiers good for? In line with the original motivation to alleviate the \emph{opacity} of learned representations, work using probing classifiers has characterized them along a range of fine-grained properties. 
However, we have discussed several reservations regarding which insights can be drawn from a probing classifier experiment. 
Absolute claims about representation \emph{quality} seem difficult to make. 
Yet recent improvements to the framework, such as better controls and metrics, allow us to make relative claims and answer questions like how \emph{extractable} a property is from a representation.  
And causal approaches (\Cref{sec:causal}) may reveal which properties are \emph{used} by the original model.

One might hope that probing classifier experiments would suggest ways to improve the quality of the probed model or to direct it to be better tuned to some use or task. Presently, there are few such successful examples. For instance, {earlier results showing that lower layers in language models focus on local phenomena while higher layers focus on global ones (using probing classifiers and other methods) motivated \citet{cao-etal-2020-deformer} to decouple a question-answering model, such that lower layers process the question and the passage independently and higher layers process them jointly. 
An analysis of redundancy in language models (again using probing classifiers and other methods) motivated an efficient transfer-learning procedure \cite{dalvi-etal-2020-analyzing}. 
An analysis of phonetic information in layers of a speech recognition systems \cite{NIPS2017_b069b341} partly motivated \citet{krishna2018hierarchical} to propose multi-task learning with phonetic supervision on intermediate layers.  
 \citet{belinkov-etal-2020-linguistic} discuss how their probing experiments can guide the selection of which machine translation models to use when translating specific languages. 
Finally, when considering using the representations for some downstream task, probing experiments can indicate which information is encoded, or can easily be extracted, from these representations.

To conclude, our critical review of the probing classifiers framework reveals that it is more complicated than may seem. 
When designing a probing classifier experiment, we advise researchers to take the various controls and alternative measures into account. Naturally, one should clearly define the original task/dataset/model and the probing task/dataset/classifier. It is important to set upper and lower bounds, and to consider proper controls, via either control tasks (for word-level properties) or datasets (for sentence-level properties). Depending on goals, one may want to measure the probe's complexity (if ease of extractability is in question), report the accuracy--complexity trade-off (when designing new probes), or perform an intervention (to measure usage of information by the original model). When possible, using parameter-free probes may circumvent some of the challenges with parameterized probes. 
We do not argue that every study must perform all the various controls and report all the alternative measures summarized here. 
However, future work seeking to use probing classifiers would do well to take into account the complexity of the framework, its apparent shortcomings, and available advances.

\begin{acknowledgments}
This research was supported by the ISRAEL SCIENCE FOUNDATION (grant No. 448/20) and by an Azrieli Foundation Early Career Faculty Fellowship.
\end{acknowledgments}

\starttwocolumn
\bibliography{compling_style}

\begin{thebibliography}{59}
\expandafter\ifx\csname natexlab\endcsname\relax\def\natexlab#1{#1}\fi

\bibitem[{Adi et~al.(2016)Adi, Kermany, Belinkov, Lavi, and
  Goldberg}]{DBLP:journals/corr/AdiKBLG16}
Adi, Yossi, Einat Kermany, Yonatan Belinkov, Ofer Lavi, and Yoav Goldberg.
  2016.
\newblock Fine-grained analysis of sentence embeddings using auxiliary
  prediction tasks.
\newblock \emph{CoRR}, abs/1608.04207.

\bibitem[{Adi et~al.(2017)Adi, Kermany, Belinkov, Lavi, and
  Goldberg}]{adi:2017:ICLR}
Adi, Yossi, Einat Kermany, Yonatan Belinkov, Ofer Lavi, and Yoav Goldberg.
  2017.
\newblock Fine-grained analysis of sentence embeddings using auxiliary
  prediction tasks.
\newblock In \emph{International Conference on Learning Representations
  (ICLR)}.

\bibitem[{Alain and Bengio(2016)}]{alain2016understanding}
Alain, Guillaume and Yoshua Bengio. 2016.
\newblock Understanding intermediate layers using linear classifier probes.
\newblock \emph{arXiv preprint arXiv:1610.01644v3}.

\bibitem[{Bau et~al.(2019)Bau, Belinkov, Sajjad, Durrani, Dalvi, and
  Glass}]{bau2018identifying}
Bau, Anthony, Yonatan Belinkov, Hassan Sajjad, Nadir Durrani, Fahim Dalvi, and
  James Glass. 2019.
\newblock Identifying and controlling important neurons in neural machine
  translation.
\newblock In \emph{International Conference on Learning Representations}.

\bibitem[{Belinkov(2018)}]{belinkov:2018:phdthesis}
Belinkov, Yonatan. 2018.
\newblock \emph{On Internal Language Representations in Deep Learning: An
  Analysis of Machine Translation and Speech Recognition}.
\newblock Ph.D. thesis, Massachusetts Institute of Technology.

\bibitem[{Belinkov et~al.(2017{\natexlab{a}})Belinkov, Durrani, Dalvi, Sajjad,
  and Glass}]{belinkov-etal-2017-neural}
Belinkov, Yonatan, Nadir Durrani, Fahim Dalvi, Hassan Sajjad, and James Glass.
  2017{\natexlab{a}}.
\newblock What do neural machine translation models learn about morphology?
\newblock In \emph{Proceedings of the 55th Annual Meeting of the Association
  for Computational Linguistics (Volume 1: Long Papers)}, pages 861--872,
  Association for Computational Linguistics, Vancouver, Canada.

\bibitem[{Belinkov et~al.(2020)Belinkov, Durrani, Dalvi, Sajjad, and
  Glass}]{belinkov-etal-2020-linguistic}
Belinkov, Yonatan, Nadir Durrani, Fahim Dalvi, Hassan Sajjad, and James Glass.
  2020.
\newblock On the linguistic representational power of neural machine
  translation models.
\newblock \emph{Computational Linguistics}, 46(1):1--52.

\bibitem[{Belinkov, Gehrmann, and
  Pavlick(2020)}]{belinkov-etal-2020-interpretability}
Belinkov, Yonatan, Sebastian Gehrmann, and Ellie Pavlick. 2020.
\newblock Interpretability and analysis in neural {NLP}.
\newblock In \emph{Proceedings of the 58th Annual Meeting of the Association
  for Computational Linguistics: Tutorial Abstracts}, pages 1--5, Association
  for Computational Linguistics, Online.

\bibitem[{Belinkov and Glass(2017)}]{NIPS2017_b069b341}
Belinkov, Yonatan and James Glass. 2017.
\newblock Analyzing hidden representations in end-to-end automatic speech
  recognition systems.
\newblock In \emph{Advances in Neural Information Processing Systems},
  volume~30, pages 2441--2451, Curran Associates, Inc.

\bibitem[{Belinkov and Glass(2019)}]{belinkov-glass-2019-analysis}
Belinkov, Yonatan and James Glass. 2019.
\newblock Analysis methods in neural language processing: A survey.
\newblock \emph{Transactions of the Association for Computational Linguistics},
  7:49--72.

\bibitem[{Belinkov et~al.(2017{\natexlab{b}})Belinkov, M{\`a}rquez, Sajjad,
  Durrani, Dalvi, and Glass}]{belinkov-etal-2017-evaluating}
Belinkov, Yonatan, Llu{\'\i}s M{\`a}rquez, Hassan Sajjad, Nadir Durrani, Fahim
  Dalvi, and James Glass. 2017{\natexlab{b}}.
\newblock Evaluating layers of representation in neural machine translation on
  part-of-speech and semantic tagging tasks.
\newblock In \emph{Proceedings of the Eighth International Joint Conference on
  Natural Language Processing (Volume 1: Long Papers)}, pages 1--10, Asian
  Federation of Natural Language Processing, Taipei, Taiwan.

\bibitem[{Cao et~al.(2020)Cao, Trivedi, Balasubramanian, and
  Balasubramanian}]{cao-etal-2020-deformer}
Cao, Qingqing, Harsh Trivedi, Aruna Balasubramanian, and Niranjan
  Balasubramanian. 2020.
\newblock {D}e{F}ormer: Decomposing pre-trained transformers for faster
  question answering.
\newblock In \emph{Proceedings of the 58th Annual Meeting of the Association
  for Computational Linguistics}, pages 4487--4497, Association for
  Computational Linguistics, Online.

\bibitem[{Cao, Sanh, and Rush(2021)}]{cao2021low}
Cao, Steven, Victor Sanh, and Alexander Rush. 2021.
\newblock Low-complexity probing via finding subnetworks.
\newblock In \emph{Proceedings of the 2021 Conference of the North American
  Chapter of the Association for Computational Linguistics: Human Language
  Technologies}, pages 960--966, Association for Computational Linguistics,
  Online.

\bibitem[{Chrupa{\l}a and Alishahi(2019)}]{chrupala-alishahi-2019-correlating}
Chrupa{\l}a, Grzegorz and Afra Alishahi. 2019.
\newblock Correlating neural and symbolic representations of language.
\newblock In \emph{Proceedings of the 57th Annual Meeting of the Association
  for Computational Linguistics}, pages 2952--2962, Association for
  Computational Linguistics, Florence, Italy.

\bibitem[{Chrupa{\l}a, Higy, and Alishahi(2020)}]{chrupala-etal-2020-analyzing}
Chrupa{\l}a, Grzegorz, Bertrand Higy, and Afra Alishahi. 2020.
\newblock Analyzing analytical methods: The case of phonology in neural models
  of spoken language.
\newblock In \emph{Proceedings of the 58th Annual Meeting of the Association
  for Computational Linguistics}, pages 4146--4156, Association for
  Computational Linguistics, Online.

\bibitem[{CHU(1965)}]{10030090917}
CHU, Y. 1965.
\newblock On the shortest arborescence of a directed graph.
\newblock \emph{Science Sinica}, 14:1396--1400.

\bibitem[{Clark et~al.(2019)Clark, Khandelwal, Levy, and
  Manning}]{clark-etal-2019-bert}
Clark, Kevin, Urvashi Khandelwal, Omer Levy, and Christopher~D. Manning. 2019.
\newblock What does {BERT} look at? an analysis of {BERT}{'}s attention.
\newblock In \emph{Proceedings of the 2019 ACL Workshop BlackboxNLP: Analyzing
  and Interpreting Neural Networks for NLP}, pages 276--286, Association for
  Computational Linguistics, Florence, Italy.

\bibitem[{Conneau et~al.(2018)Conneau, Kruszewski, Lample, Barrault, and
  Baroni}]{conneau-etal-2018-cram}
Conneau, Alexis, German Kruszewski, Guillaume Lample, Lo{\"\i}c Barrault, and
  Marco Baroni. 2018.
\newblock What you can cram into a single {\$}{\&}!{\#}* vector: Probing
  sentence embeddings for linguistic properties.
\newblock In \emph{Proceedings of the 56th Annual Meeting of the Association
  for Computational Linguistics (Volume 1: Long Papers)}, pages 2126--2136,
  Association for Computational Linguistics, Melbourne, Australia.

\bibitem[{Dalvi et~al.(2020)Dalvi, Sajjad, Durrani, and
  Belinkov}]{dalvi-etal-2020-analyzing}
Dalvi, Fahim, Hassan Sajjad, Nadir Durrani, and Yonatan Belinkov. 2020.
\newblock Analyzing redundancy in pretrained transformer models.
\newblock In \emph{Proceedings of the 2020 Conference on Empirical Methods in
  Natural Language Processing (EMNLP)}, pages 4908--4926, Association for
  Computational Linguistics, Online.

\bibitem[{Danilevsky et~al.(2020)Danilevsky, Qian, Aharonov, Katsis, Kawas, and
  Sen}]{danilevsky-etal-2020-survey}
Danilevsky, Marina, Kun Qian, Ranit Aharonov, Yannis Katsis, Ban Kawas, and
  Prithviraj Sen. 2020.
\newblock A survey of the state of explainable {AI} for natural language
  processing.
\newblock In \emph{Proceedings of the 1st Conference of the Asia-Pacific
  Chapter of the Association for Computational Linguistics and the 10th
  International Joint Conference on Natural Language Processing}, pages
  447--459, Association for Computational Linguistics, Suzhou, China.

\bibitem[{Devlin et~al.(2019)Devlin, Chang, Lee, and
  Toutanova}]{devlin-etal-2019-bert}
Devlin, Jacob, Ming-Wei Chang, Kenton Lee, and Kristina Toutanova. 2019.
\newblock {BERT}: Pre-training of deep bidirectional transformers for language
  understanding.
\newblock In \emph{Proceedings of the 2019 Conference of the North {A}merican
  Chapter of the Association for Computational Linguistics: Human Language
  Technologies, Volume 1 (Long and Short Papers)}, pages 4171--4186,
  Association for Computational Linguistics, Minneapolis, Minnesota.

\bibitem[{Edmonds(1967)}]{edmonds1967optimum}
Edmonds, Jack. 1967.
\newblock Optimum branchings.
\newblock \emph{Journal of Research of the national Bureau of Standards B},
  71(4):233--240.

\bibitem[{Elazar et~al.(2021)Elazar, Ravfogel, Jacovi, and
  Goldberg}]{elazar2020amnesic}
Elazar, Yanai, Shauli Ravfogel, Alon Jacovi, and Yoav Goldberg. 2021.
\newblock Amnesic probing: Behavioral explanation with amnesic counterfactuals.
\newblock \emph{Transactions of the Association for Computational Linguistics},
  9(0):160--175.

\bibitem[{Ettinger, Elgohary, and Resnik(2016)}]{ettinger-etal-2016-probing}
Ettinger, Allyson, Ahmed Elgohary, and Philip Resnik. 2016.
\newblock Probing for semantic evidence of composition by means of simple
  classification tasks.
\newblock In \emph{Proceedings of the 1st Workshop on Evaluating Vector-Space
  Representations for {NLP}}, pages 134--139, Association for Computational
  Linguistics, Berlin, Germany.

\bibitem[{Feder et~al.(2021)Feder, Oved, Shalit, and
  Reichart}]{feder2020causalm}
Feder, Amir, Nadav Oved, Uri Shalit, and Roi Reichart. 2021.
\newblock {CausaLM: Causal Model Explanation Through Counterfactual Language
  Models}.
\newblock \emph{Computational Linguistics}, 47(2):333--386.

\bibitem[{Giulianelli et~al.(2018)Giulianelli, Harding, Mohnert, Hupkes, and
  Zuidema}]{giulianelli-etal-2018-hood}
Giulianelli, Mario, Jack Harding, Florian Mohnert, Dieuwke Hupkes, and Willem
  Zuidema. 2018.
\newblock Under the hood: Using diagnostic classifiers to investigate and
  improve how language models track agreement information.
\newblock In \emph{Proceedings of the 2018 {EMNLP} Workshop {B}lackbox{NLP}:
  Analyzing and Interpreting Neural Networks for {NLP}}, pages 240--248,
  Association for Computational Linguistics, Brussels, Belgium.

\bibitem[{Gupta et~al.(2015)Gupta, Boleda, Baroni, and
  Pad{\'o}}]{gupta-etal-2015-distributional}
Gupta, Abhijeet, Gemma Boleda, Marco Baroni, and Sebastian Pad{\'o}. 2015.
\newblock Distributional vectors encode referential attributes.
\newblock In \emph{Proceedings of the 2015 Conference on Empirical Methods in
  Natural Language Processing}, pages 12--21, Association for Computational
  Linguistics, Lisbon, Portugal.

\bibitem[{Hall~Maudslay et~al.(2020)Hall~Maudslay, Valvoda, Pimentel, Williams,
  and Cotterell}]{hall-maudslay-etal-2020-tale}
Hall~Maudslay, Rowan, Josef Valvoda, Tiago Pimentel, Adina Williams, and Ryan
  Cotterell. 2020.
\newblock A tale of a probe and a parser.
\newblock In \emph{Proceedings of the 58th Annual Meeting of the Association
  for Computational Linguistics}, pages 7389--7395, Association for
  Computational Linguistics, Online.

\bibitem[{Hewitt and Liang(2019)}]{hewitt-liang-2019-designing}
Hewitt, John and Percy Liang. 2019.
\newblock Designing and interpreting probes with control tasks.
\newblock In \emph{Proceedings of the 2019 Conference on Empirical Methods in
  Natural Language Processing and the 9th International Joint Conference on
  Natural Language Processing (EMNLP-IJCNLP)}, pages 2733--2743, Association
  for Computational Linguistics, Hong Kong, China.

\bibitem[{Htut et~al.(2019)Htut, Phang, Bordia, and Bowman}]{htut2019attention}
Htut, Phu~Mon, Jason Phang, Shikha Bordia, and Samuel~R Bowman. 2019.
\newblock Do attention heads in bert track syntactic dependencies?
\newblock \emph{arXiv preprint arXiv:1911.12246}.

\bibitem[{Hupkes, Veldhoen, and Zuidema(2018)}]{hupkes2018visualisation}
Hupkes, Dieuwke, Sara Veldhoen, and Willem Zuidema. 2018.
\newblock Visualisation and 'diagnostic classifiers' reveal how recurrent and
  recursive neural networks process hierarchical structure.
\newblock \emph{Journal of Artificial Intelligence Research}, 61:907--926.

\bibitem[{K{\"o}hn(2015)}]{kohn-2015-whats}
K{\"o}hn, Arne. 2015.
\newblock What{'}s in an embedding? analyzing word embeddings through
  multilingual evaluation.
\newblock In \emph{Proceedings of the 2015 Conference on Empirical Methods in
  Natural Language Processing}, pages 2067--2073, Association for Computational
  Linguistics, Lisbon, Portugal.

\bibitem[{Kriegeskorte, Mur, and Bandettini(2008)}]{10.3389/neuro.06.004.2008}
Kriegeskorte, Nikolaus, Marieke Mur, and Peter Bandettini. 2008.
\newblock Representational similarity analysis - connecting the branches of
  systems neuroscience.
\newblock \emph{Frontiers in Systems Neuroscience}, 2:4.

\bibitem[{Krishna, Toshniwal, and Livescu(2019)}]{krishna2018hierarchical}
Krishna, Kalpesh, Shubham Toshniwal, and Karen Livescu. 2019.
\newblock Hierarchical multitask learning for {CTC}-based speech recognition.
\newblock \emph{arXiv preprint arXiv:1807.06234}.

\bibitem[{Lakretz et~al.(2019)Lakretz, Kruszewski, Desbordes, Hupkes, Dehaene,
  and Baroni}]{lakretz-etal-2019-emergence}
Lakretz, Yair, German Kruszewski, Theo Desbordes, Dieuwke Hupkes, Stanislas
  Dehaene, and Marco Baroni. 2019.
\newblock The emergence of number and syntax units in {LSTM} language models.
\newblock In \emph{Proceedings of the 2019 Conference of the North {A}merican
  Chapter of the Association for Computational Linguistics: Human Language
  Technologies, Volume 1 (Long and Short Papers)}, pages 11--20, Association
  for Computational Linguistics, Minneapolis, Minnesota.

\bibitem[{Lepori and McCoy(2020)}]{lepori-mccoy-2020-picking}
Lepori, Michael and R.~Thomas McCoy. 2020.
\newblock Picking {BERT}{'}s brain: Probing for linguistic dependencies in
  contextualized embeddings using representational similarity analysis.
\newblock In \emph{Proceedings of the 28th International Conference on
  Computational Linguistics}, pages 3637--3651, International Committee on
  Computational Linguistics, Barcelona, Spain (Online).

\bibitem[{Liu et~al.(2019)Liu, Gardner, Belinkov, Peters, and
  Smith}]{liu-etal-2019-linguistic}
Liu, Nelson~F., Matt Gardner, Yonatan Belinkov, Matthew~E. Peters, and Noah~A.
  Smith. 2019.
\newblock Linguistic knowledge and transferability of contextual
  representations.
\newblock In \emph{Proceedings of the 2019 Conference of the North {A}merican
  Chapter of the Association for Computational Linguistics: Human Language
  Technologies, Volume 1 (Long and Short Papers)}, pages 1073--1094,
  Association for Computational Linguistics, Minneapolis, Minnesota.

\bibitem[{Lovering et~al.(2021)Lovering, Jha, Linzen, and
  Pavlick}]{lovering2021predicting}
Lovering, Charles, Rohan Jha, Tal Linzen, and Ellie Pavlick. 2021.
\newblock Predicting inductive biases of pre-trained models.
\newblock In \emph{International Conference on Learning Representations}.

\bibitem[{Mare{\v{c}}ek and Rosa(2019)}]{marecek-rosa-2019-balustrades}
Mare{\v{c}}ek, David and Rudolf Rosa. 2019.
\newblock From balustrades to pierre vinken: Looking for syntax in transformer
  self-attentions.
\newblock In \emph{Proceedings of the 2019 ACL Workshop BlackboxNLP: Analyzing
  and Interpreting Neural Networks for NLP}, pages 263--275, Association for
  Computational Linguistics, Florence, Italy.

\bibitem[{Michael, Botha, and Tenney(2020)}]{michael-etal-2020-asking}
Michael, Julian, Jan~A. Botha, and Ian Tenney. 2020.
\newblock Asking without telling: Exploring latent ontologies in contextual
  representations.
\newblock In \emph{Proceedings of the 2020 Conference on Empirical Methods in
  Natural Language Processing (EMNLP)}, pages 6792--6812, Association for
  Computational Linguistics, Online.

\bibitem[{Pimentel et~al.(2020{\natexlab{a}})Pimentel, Saphra, Williams, and
  Cotterell}]{pimentel-etal-2020-pareto}
Pimentel, Tiago, Naomi Saphra, Adina Williams, and Ryan Cotterell.
  2020{\natexlab{a}}.
\newblock {P}areto probing: {T}rading off accuracy for complexity.
\newblock In \emph{Proceedings of the 2020 Conference on Empirical Methods in
  Natural Language Processing (EMNLP)}, pages 3138--3153, Association for
  Computational Linguistics, Online.

\bibitem[{Pimentel et~al.(2020{\natexlab{b}})Pimentel, Valvoda, Hall~Maudslay,
  Zmigrod, Williams, and Cotterell}]{pimentel-etal-2020-information}
Pimentel, Tiago, Josef Valvoda, Rowan Hall~Maudslay, Ran Zmigrod, Adina
  Williams, and Ryan Cotterell. 2020{\natexlab{b}}.
\newblock Information-theoretic probing for linguistic structure.
\newblock In \emph{Proceedings of the 58th Annual Meeting of the Association
  for Computational Linguistics}, pages 4609--4622, Association for
  Computational Linguistics, Online.

\bibitem[{Raganato and Tiedemann(2018)}]{raganato-tiedemann-2018-analysis}
Raganato, Alessandro and J{\"o}rg Tiedemann. 2018.
\newblock An analysis of encoder representations in transformer-based machine
  translation.
\newblock In \emph{Proceedings of the 2018 {EMNLP} Workshop {B}lackbox{NLP}:
  Analyzing and Interpreting Neural Networks for {NLP}}, pages 287--297,
  Association for Computational Linguistics, Brussels, Belgium.

\bibitem[{Ravichander, Belinkov, and Hovy(2021)}]{ravichander:2021:eacl}
Ravichander, Abhilasha, Yonatan Belinkov, and Eduard Hovy. 2021.
\newblock Probing the probing paradigm: Does probing accuracy entail task
  relevance?
\newblock In \emph{Proceedings of the 16th Conference of the European Chapter
  of the Association for Computational Linguistics: Main Volume}, pages
  3363--3377, Association for Computational Linguistics, Online.

\bibitem[{Rogers, Kovaleva, and Rumshisky(2020)}]{rogers-etal-2020-primer}
Rogers, Anna, Olga Kovaleva, and Anna Rumshisky. 2020.
\newblock A primer in {BERT}ology: What we know about how {BERT} works.
\newblock \emph{Transactions of the Association for Computational Linguistics},
  8:842--866.

\bibitem[{Shi, Padhi, and Knight(2016)}]{shi-etal-2016-string}
Shi, Xing, Inkit Padhi, and Kevin Knight. 2016.
\newblock Does string-based neural {MT} learn source syntax?
\newblock In \emph{Proceedings of the 2016 Conference on Empirical Methods in
  Natural Language Processing}, pages 1526--1534, Association for Computational
  Linguistics, Austin, Texas.

\bibitem[{Tamkin et~al.(2020)Tamkin, Singh, Giovanardi, and
  Goodman}]{tamkin-etal-2020-investigating}
Tamkin, Alex, Trisha Singh, Davide Giovanardi, and Noah Goodman. 2020.
\newblock Investigating transferability in pretrained language models.
\newblock In \emph{Findings of the Association for Computational Linguistics:
  EMNLP 2020}, pages 1393--1401, Association for Computational Linguistics,
  Online.

\bibitem[{Tenney et~al.(2019)Tenney, Xia, Chen, Wang, Poliak, McCoy, Kim,
  Durme, Bowman, Das, and Pavlick}]{tenney2018what}
Tenney, Ian, Patrick Xia, Berlin Chen, Alex Wang, Adam Poliak, R~Thomas McCoy,
  Najoung Kim, Benjamin~Van Durme, Sam Bowman, Dipanjan Das, and Ellie Pavlick.
  2019.
\newblock What do you learn from context? probing for sentence structure in
  contextualized word representations.
\newblock In \emph{International Conference on Learning Representations}.

\bibitem[{Tucker, Qian, and Levy(2021)}]{tucker2021modified}
Tucker, Mycal, Peng Qian, and Roger Levy. 2021.
\newblock What if this modified that? syntactic interventions with
  counterfactual embeddings.
\newblock In \emph{Findings of the Association for Computational Linguistics:
  ACL-IJCNLP 2021}, pages 862--875, Association for Computational Linguistics,
  Online.

\bibitem[{Vanmassenhove, Du, and Way(2017)}]{VanmassenhoveDuWay2017}
Vanmassenhove, Eva, Jinhua Du, and Andy Way. 2017.
\newblock Investigating `aspect' in {NMT} and {SMT}: Translating the english
  simple past and present perfect.
\newblock \emph{Computational Linguistics in the Netherlands Journal},
  7:109--128.

\bibitem[{Veldhoen, Hupkes, and Zuidema(2016)}]{veldhoen2016diagnostic}
Veldhoen, Sara, Dieuwke Hupkes, and Willem~H Zuidema. 2016.
\newblock Diagnostic classifiers revealing how neural networks process
  hierarchical structure.
\newblock In \emph{CoCo@ NIPS}.

\bibitem[{Vig et~al.(2020)Vig, Gehrmann, Belinkov, Qian, Nevo, Singer, and
  Shieber}]{vig:2020:neurips}
Vig, Jesse, Sebastian Gehrmann, Yonatan Belinkov, Sharon Qian, Daniel Nevo,
  Yaron Singer, and Stuart Shieber. 2020.
\newblock Investigating gender bias in language models using causal mediation
  analysis.
\newblock In \emph{Advances in Neural Information Processing Systems},
  volume~33, pages 12388--12401, Curran Associates, Inc.

\bibitem[{Voita and Titov(2020)}]{voita-titov-2020-information}
Voita, Elena and Ivan Titov. 2020.
\newblock Information-theoretic probing with minimum description length.
\newblock In \emph{Proceedings of the 2020 Conference on Empirical Methods in
  Natural Language Processing (EMNLP)}, pages 183--196, Association for
  Computational Linguistics, Online.

\bibitem[{Wallace, Gardner, and Singh(2020)}]{wallace-etal-2020-interpreting}
Wallace, Eric, Matt Gardner, and Sameer Singh. 2020.
\newblock Interpreting predictions of {NLP} models.
\newblock In \emph{Proceedings of the 2020 Conference on Empirical Methods in
  Natural Language Processing: Tutorial Abstracts}, pages 20--23, Association
  for Computational Linguistics, Online.

\bibitem[{Wu et~al.(2020)Wu, Chen, Kao, and Liu}]{wu-etal-2020-perturbed}
Wu, Zhiyong, Yun Chen, Ben Kao, and Qun Liu. 2020.
\newblock Perturbed masking: Parameter-free probing for analyzing and
  interpreting {BERT}.
\newblock In \emph{Proceedings of the 58th Annual Meeting of the Association
  for Computational Linguistics}, pages 4166--4176, Association for
  Computational Linguistics, Online.

\bibitem[{Zhang and Bowman(2018)}]{zhang-bowman-2018-language}
Zhang, Kelly and Samuel Bowman. 2018.
\newblock Language modeling teaches you more than translation does: Lessons
  learned through auxiliary syntactic task analysis.
\newblock In \emph{Proceedings of the 2018 {EMNLP} Workshop {B}lackbox{NLP}:
  Analyzing and Interpreting Neural Networks for {NLP}}, pages 359--361,
  Association for Computational Linguistics, Brussels, Belgium.

\bibitem[{Zhang et~al.(2021)Zhang, Warstadt, Li, and
  Bowman}]{zhang-etal-2021-need}
Zhang, Yian, Alex Warstadt, Xiaocheng Li, and Samuel~R. Bowman. 2021.
\newblock When do you need billions of words of pretraining data?
\newblock In \emph{Proceedings of the 59th Annual Meeting of the Association
  for Computational Linguistics and the 11th International Joint Conference on
  Natural Language Processing (Volume 1: Long Papers)}, pages 1112--1125,
  Association for Computational Linguistics, Online.

\bibitem[{Zhou and Srikumar(2021)}]{zhou-srikumar-2021}
Zhou, Yichu and Vivek Srikumar. 2021.
\newblock {D}irect{P}robe: Studying representations without classifiers.
\newblock In \emph{Proceedings of the 2021 Conference of the North American
  Chapter of the Association for Computational Linguistics: Human Language
  Technologies}, pages 5070--5083, Association for Computational Linguistics,
  Online.

\bibitem[{Zhu and Rudzicz(2020)}]{zhu-rudzicz-2020-information}
Zhu, Zining and Frank Rudzicz. 2020.
\newblock An information theoretic view on selecting linguistic probes.
\newblock In \emph{Proceedings of the 2020 Conference on Empirical Methods in
  Natural Language Processing (EMNLP)}, pages 9251--9262, Association for
  Computational Linguistics, Online.

\end{thebibliography}

\end{document}